\documentclass{article}

\usepackage{arxiv}

\usepackage[utf8]{inputenc} 
\usepackage[T1]{fontenc}    
\usepackage{hyperref}       
\usepackage{url}            
\usepackage{booktabs}       
\usepackage{amsfonts}       
\usepackage{nicefrac}       
\usepackage{microtype}      
\usepackage{lipsum}		
\usepackage{graphicx}
\usepackage{natbib}
\usepackage{doi}
\usepackage{url,hyperref,lineno,microtype,subcaption,ulem,soul,hyperref, color}
\usepackage{amsmath}

\title{Training for temporal sparsity in deep neural networks, application in video processing}

\author{ \href{https://orcid.org/0000-0002-2967-5090}{\includegraphics[scale=0.06]{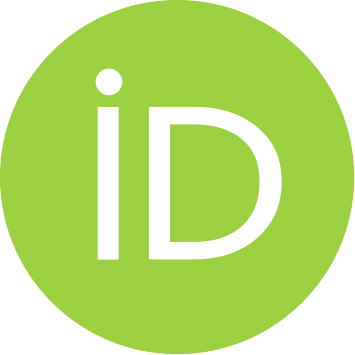}\hspace{1mm}Amirreza Yousefzadeh} \\

	\texttt{Amirreza.Yousefzadeh@imec.nl} \\
	\And
	\href{https://orcid.org/0000-0002-0949-2094}{\includegraphics[scale=0.06]{orcid.pdf}\hspace{1mm}Manolis Sifalakis}
	\\
	\texttt{Manolis.Sifalakis@imec.nl} \\
}

\renewcommand{\shorttitle}{\textit{arXiv} Template}

\hypersetup{
pdftitle={A template for the arxiv style},
pdfsubject={q-bio.NC, q-bio.QM},
pdfauthor={David S.~Hippocampus, Elias D.~Striatum},
pdfkeywords={First keyword, Second keyword, More},
}

\begin{document}
\maketitle

\begin{abstract}

Activation sparsity improves compute efficiency and resource utilization in sparsity-aware neural network accelerators. As the predominant operation in DNNs is multiply-accumulate (MAC) of activations with weights to compute inner products, skipping operations where (at least) one of the two operands is zero can make inference more efficient in terms of latency and power. Spatial sparsification of activations is a popular topic in DNN literature and several methods have already been established to bias a DNN for it (e.g. regularization, quantization, boxing, etc). On the other hand, temporal sparsity is an inherent feature of bio-inspired spiking neural networks (SNNs), which neuromorphic processing exploits for hardware efficiency. Introducing and exploiting spatio-temporal sparsity, is a topic much less explored in DNN literature, but in perfect resonance with the trend in DNN, to shift from static signal processing (e.g. image processing) to more streaming signal processing (e.g. video and audio).

Towards this goal, in this paper we introduce a new DNN layer (called Delta Activation Layer), whose sole purpose is to promote temporal sparsity of activations during training. A Delta Activation Layer casts temporal sparsity into spatial activation sparsity to be exploited when performing sparse tensor multiplications in hardware. By employing delta inference and ``the usual'' spatial sparsification heuristics during training, the resulting model learns to exploit not only spatial but also temporal activation sparsity (for a given input data distribution). One may use the Delta Activation Layer either during vanilla training or during a refinement phase. 

We have implemented Delta Activation Layer as an extension of the standard Tensoflow-Keras (2.0) library, and applied it to train deep neural networks on the Human Action Recognition (UCF101) dataset. We report an almost 3x improvement of activation sparsity, with recoverable loss of model accuracy after longer training.

For reproducibility of the results we have made available the source for the Delta Activation Layer at \url{https://github.com/msifalakis/delta_activation_layer}.

\end{abstract}

\section{Introduction}

As the depth, size, and complexity of neural network models increases, training but also inference is becoming increasingly more expensive in terms of resources to perform, often even on tailor-made accelerator platforms. Therefore, model compression and efficiently compressed inference processing has became very important in DNN literature \citep{efficient_DNN_survey}; and over years, several strategies and solutions have been proposed for reducing the memory and compute requirements of DNN models. 

By training with sparsity penalties, and/or employing clever quantization, and network pruning heuristics, e.g. \citep{Han2016DeepCC} \citep{sparseDNN2019}, it is possible to reduce the network size, so as to consume less memory and perform less operations, with often unsubstantial loss (or even improvement) of accuracy. Likewise, by employing dual-training of teacher and student models\citep{hinton2015distilling}, it has been shown possible to produce very compact and stable (student) inference models that even generalize better (than the teacher).

In all these approaches and strategies, sparsity is paramount as sparse tensors are not only easier to  store and access in memory, but can also be (under conditions) more efficient to process. In practical reality, performing sparse operations on parallelizing accelerators often accounts a little extra effort per operation due to the fact that the resulting irregularity of computation cannot be optimally distributed across the parallel cores. In this case if the sparsity level is not sufficiently high, executing sparse operations in off-the-shelf accelerators will even be less efficient\footnote{For example sparsity less than 98\% in NVIDIA V100 GPU when using NVIDIA cuSPARSE library results in less processing throughput than dense execution \citep{sparseMM_GPU}}. 


Interestingly, as the competition for tera-operations (TOPS) per Watt in the DNN accelerator architectures has become less relevant over time \citep{sze2020evaluate}, the focus of academic research in DNN accelerator and architectures has shifted towards sparsity exploitation in memory and compute operations \citep{albericio2016cnvlutin} \citep{han2016eie} \citep{parashar2017scnn} \citep{aimar2018nullhop} \citep{kepner2020graphchallenge}. This resulted in the new terminology of “effective TOPS"\footnote{Effective TOPS refers to the equivalent amount of TOPS when the sparsity is not taken into account. For an accelerator that can exploit sparsity, effective TOPS is higher than its actual TOPS.}. Besides academic research, several start-ups are developing commercial versions of sparsity exploiting neural network inference processors \citep{NeuronFlow} \citep{brainchip_akida_linley2019}. Finally, recently the new NVIDIA GPU A100 architecture was released with features to exploit fine-grained sparsity in deep learning networks, which doubles the throughput of Tensor Core operations."\citep{NVIDIA_ampere}.  

\begin{figure}[h!]
\begin{center}
\includegraphics[width=15cm]{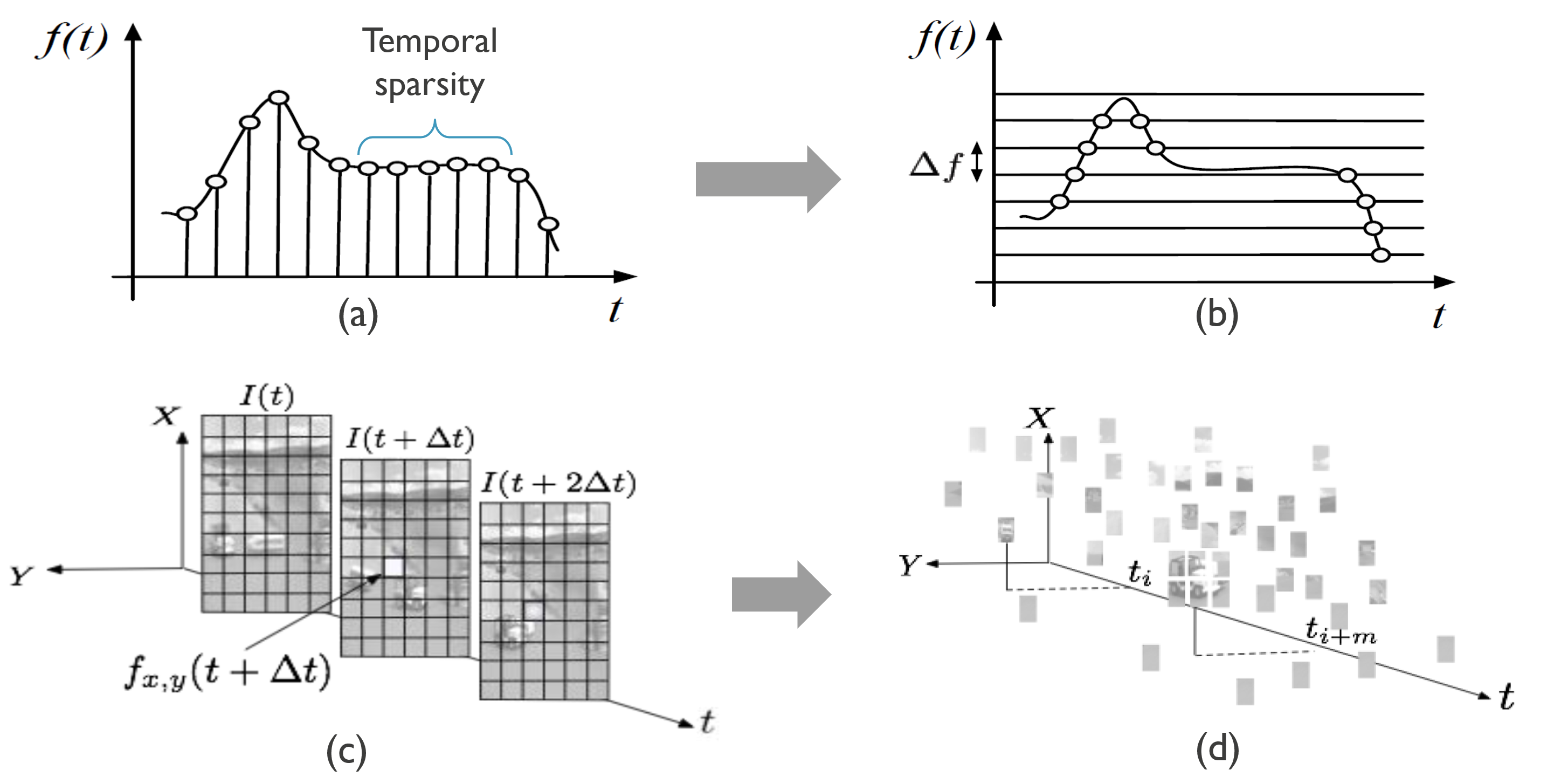}
\end{center}
\caption{(a) Concept of temporal sparsity in a signal, (b) temporally sparse processing, (c) Sequence of 2D images in time from a video, (d) Temporally sparse processing of video frames. Images are taken from \citep{temporal_sparsity_figure}}\label{fig:temp_sparsity}
\end{figure}

Unlike structural (weight) sparsity and spatial (activation) sparsity, discussed so far, which are hot topics in DNN literature \citep{wen2016learning}, there is also temporal activation sparsity, which is less explored in the context of DNN, yet rather popular theme in signal processing (compressed sensing), and neuromorphic computing. By definition, temporal sparsity exists in a signal which is not changing over time. Fig. \ref{fig:temp_sparsity} (a) shows evolution of a signal $f(t)$ over time. For processing this signal in DNN, it is sampled in a fixed periodic time-steps and every sample is processed separately. Since some of the samples in time may be redundant, their processing can be safely skipped, as shown in Fig. \ref{fig:temp_sparsity} (b). Fig. \ref{fig:temp_sparsity} (c) and (d) shows the same concept in a two dimensional signal (sequence of video frames). And since processing of temporal signals maps to the same linear algebra operations as typical of DNNs, temporal sparsity can also lead to skip processing of zeros in sparse matrix operations sparing redundant operations. Unsurprisingly as streaming and sequence data applications (for example real-time video/audio processing) have become mainstream in DNNs, interest in exploiting temporal sparsity is relevant and slowly growing. The questions we would like to answers in this paper are 1) how it is possible to exploit temporal sparsity in a DNN, 2) how much sparsity can be added when exploiting temporal sparsity on top of the spatial sparsity and 3) what are the consequences of this technique (disadvantages).      

In compressed sensing, temporal sparsity is used in signal compression for efficient use of memory and communication bandwidth. As natural signals are inherently sparse in time (low information density), most video/audio stream recordings contain too much redundant information. Video/Audio compression algorithms exploiting the spatio-temporal sparsity in these signals can easily achieve a compression factor of 100x to 1000x \citep{richardson2004h}. 

More pertinent to the field of deep learning and machine learning (ML), neuromorphic computing exploits spatio-temporal sparsity as one of the key biology motivated principles of neural processing, for engineering hardware efficient algorithms and scalable, energy efficient, processing architectures. Unlike DNN accelerators, a design tenet of Spiking Neural Network accelerators \citep{SpiNNaker} \citep{LOIHI} \citep{TrueNorth} has always been to perform efficient sparse operations; as a consequence of the very degree of temporal sparsity in the activations of spiking neurons.

Motivated by neuromorphic compute-platform engineering, we present here a simplified methodology for inducing temporal sparsity to any DNN, by means of a new activation layer (called Delta Activation Layer), that can be introduced in a DNN at any phase (training, refinement, or inference only). A Delta Activation Layer contains rather simple stateful neurons that remember past activations and quantize-propagate only changes in activations over time. While it functions similar to delta-inference~\citep{neil2017delta} or sigma-delta networks~\citep{o2017sigma}, there is however no threshold~\citep{neil2017delta} or PID parameters~\citep{o2017sigma} to be learned. Instead a generic activation quantization method is used that can easily integrate most existing or emerging quantization methods, e.g. \citep{yang2020training} \citep{jin2020adabits}. This extensibility is motivated by the desire to choose quantization method based on the accuracy trade-off it provides. During training of a DNN, the Delta Activation Layer adds sparsity penalty to the overall cost function of the DNN together with a proper gradient to automatically optimize the quantization level for the best sparsity-accuracy balance. In this sense, the attained accuracy is equivalent to that of a quantized DNN. This Delta Activation Layer can be introduced in all or some of the DNN layers as deemed beneficial.

In the remaining of this paper, we explain the Delta Activation Layer in more detail and present the results of experiments with ResNet-50 and the UCF101 dataset. We also relate this work to previous literature and discuss current limitations of the Delta Activation Layer and the future directions.

\subsection{Related Work}

Arguably, exploitation of (spatio-)temporal activation sparsity in neural network processing has origins in neuromorphic research \citep{linares2006spike}, and more recently it has become a hot-topic in mainstream DNN literature\footnote{Albeit exploitation of structural (aka weight) sparsity for network pruning is also very old topic in ANN literature \citep{lecun1989optimal}}. 

In \citep{wu2018compressed}, the authors defend the use of compressed videos (using H.264, HVEC, etc.) directly as inputs to a DNN, by conjecturing that the higher information density of compressed video frames compared to ``raw'' frames, makes training easier and inference faster. The type of video compression  they consider account for both spatial sparsity (in base I-frames) as well as temporal sparsity (optical flow in P-frames). While effective, this method only considers temporal sparsity in the input layer (P-frames of compressed video). In addition to that our results show that temporal sparsity can also be very high and beneficial when going deeper into the neural network, where higher-level features are extracted.

In \citep{buckler2018eva2}, the authors propose a novel DNN inference algorithm (AMS) alongside a DNN accelerator optimized for video-based inference. Inspired by video compression methods the algorithm skips the processing of invariant information across input frames (temporal sparsity), and also largely reduces the processing of motion-related features in the input; by using optical flow information from P-frames to do motion compensation directly on the activation maps of previous frames. In this way large parts of the neural network ``hybernates`` (as opposed of computing new activation maps for every frame) for more of the inputs. The AMS algorithm is integratable in any video-processing CNN model/pipeline through an ``extension'' DNN accelerator on top of any existing CNN accelerator, and improves efficiency by a factor of two. This is analogous to the functioning of our Delta Activation Layer. Also similar to our method, AMS is required to keep track of the neurons' activation state to exploit temporal sparsity.



In delta networks \citep{neil2017delta}, which are motivated by neuromorphic engineering in RNNs, and sigma-delta quantized neural networks \citep{o2017sigma}, which relate to \emph{herding} \citep{welling2009herding} in Markov random fields, each neuron only processes \textit{input signal changes} between successive timesteps, thus exploiting temporal sparsity at any layer depth. In delta-networks \citep{neil2017delta} a signal change is propagated downstream only when a certain threshold is exceeded (thereby encouraging further sparsity). By analogy in sigma-delta networks \citep{o2017sigma} a signal change is quantized to produce a stochastically discretized binary signal (spike train), which is temporally sparse. In \citep{cbinfer} the authors applied delta network inference on pre-trained CNNs with videos recorded from a fixed camera (video surveillance). As the temporal sparsity is very high in these video frames, the presented results show a considerable speedup even with the off-the-shelf GPUs, without a sparsity-aware compute architecture. And likewise in \citep{gao2018delta} the authors report a substantial inference speed-ups and power-efficiency improvements on RNN FPGA-based accelerators by exploiting temporal sparsity in delta-networks.

An issue when it comes to previous works in delta network inference is the accumulation of errors over time due to the use of a threshold (level-crossing), deeper in the network, which can lead to drift in the approximation of the activation signal over time (and hurts the accuracy). Therefore in practice, it is required to reset the state of all the neurons periodically. In sigma-delta networks a similar problem is addressed by employing a form of PID control to ``regulate'' the hysteresis \citep{o'connor2018temporally}. Another ``nuissance'' is the optimization of hyper-parameters (threshold value, step-size, ...) through an additional optimization process \citep{yousefzadeh2019asynchronous} \citep{khoei2020sparnet} \citep{cbinfer} after and outside the training loop.


While using similar concept, the function provided by the proposed Delta Activation Layer differs from both typical delta networks and sigma-delta networks. During training, the Delta Activation Layer only quantizes the activation values and therefore acts as a conventional quantization layer. The quantization step size, however, is dynamic and trainable, and is optimized by using a temporal sparsity penalty. A bigger (coarser) step-size increases the temporal sparsity but tend to reduce the accuracy. The training optimizer tries to find an optimal set of parameters, including step-size, that trade the highest possible sparsity (minimizing the sparsity penalty) and highest possible accuracy (minimizing the accuracy loss).

Our training process results in a conventional quantized DNN. During inference, sigma and delta operations are introduced at each layer's quantized output without using threshold. As the activations are already quantized with an optimum quantization level, the delta operation leads to a particularly sparse signal that merely enumerates quantizer step changes. By extension to the simpler approach in \citep{cbinfer}, which also operates on video data, our experiments show temporal sparsity can also be very high for recordings with moving cameras, since in deeper layers of the neural network, where higher-level features are extracted, movement of the camera does not introduce considerable changes.


In \citep{chen20193d} the authors demonstrated how quantization results in sparsity improvements for both temporal and spatial dimensions (with application to 3D CNNs). By contrast to this work, here we train the activation quantization level per each neuron, channel, or layer for the highest possible temporal sparsity. Potentially seen as a disadvantage is that this introduces more parameterization and requires more fine-tuning epochs. We discuss this further in section \ref{sec:discussion}.

An important caveat of the nature of the delta operation (including the work presented here), is the requirement for stateful neurons deeper in a DNN, so as to keep account of past activations. Algorithmically and in software this is a minor cost as this state is rather simple and most of the times well compressible, however in terms of a hardware accelerator implementation, this cost might not be negligible. We show in the section \ref{sec:discussion} that using full Delta Activation Layers in ResNet-50 results in 40\% increase in the usage of memory. We suggest partial use of stateful layers (where it is more optimized) to partially alleviate this problem. We believe the ultimate solution for memory limitation is advancement in technology which is discussed in section \ref{sec:discussion}.

\section{Methods} \label{sec:method}

\subsection{Delta inference}

In a general DNN, the output signal of a layer is related to it's inputs and weights with the following equations:




\begin{equation} \label{eq:neuron_state}
Z(t) = g(W, X(t)) + B
\end{equation}

\begin{equation} \label{eq:activation_fun}
O(t) = f(Z(t))
\end{equation}

\noindent where $W$ and $B$ are the weight and bias tensors(trainable parameters), $X(t)$ is the input tensor in discrete time $(t)$ \footnote{$t$ is the algorithm time-step ($t>0$), for example can be the frame number in a frame-based system} and $g$ is a linear function of the DNN layer inputs (for example dense layer matrix product, strided convolution over a region of the input, average pooling, or a combination of them). $O(t)$ is the output tensor in time $t$ and $f(.)$ is a non-linear activation function. $Z(t)$ is an intermediate variable which we can call it ``neuron state''.

One can typically introduce temporal sparsity by using the first order difference, as follows:

\begin{equation}\label{eq:layer} 
\textit{Linear operations:     }\Delta Z(t) = g(W, \Delta X(t))
\end{equation}

\begin{equation}\label{eq:sigma_nn} 
\textit{Integration(Sigma):     }Z(t) = \sum_{i=1}^{t}{\Delta Z(i)} + B = \Delta Z(t) + Z(t-1) \textit{ where } Z(0)=B 
\end{equation}

\begin{equation}\label{eq:delta_nn}
\textit{Differentiation (Delta):     }\Delta O(t) = O(t) - O(t-1) = f(Z(t)) - f(Z(t-1)) \textit{ where } f(Z(0))=0 
\end{equation}

Eq. \ref{eq:sigma_nn} and \ref{eq:delta_nn} shows the foundation of delta inference algorithms and sigma-delta networks. In Eq. \ref{eq:layer}, rather than processing input tensor ($X$) directly, only changes are processed. This is possible because $g(.)$ is a linear function. When temporal sparsity is very high, we expect $\Delta X(t)$ (and $\Delta O(t)$) to be sparse tensors, which immediately leverage zero-skipping operations in Eq. \ref{eq:layer}.  

One difference between the delta inference and the normal inference is the use of bias. In delta inference bias is only used to initialized the `neuron states' in Eq. \ref{eq:sigma_nn}. However, since bias tensors do not change over time, their delta is zero and is factored out from Eq. \ref{eq:layer}.

As neuron state is integrating all the inputs over time, $Z(t)$ in Eq. \ref{eq:neuron_state} and \ref{eq:sigma_nn} are always equal, and so long as the input is the same, both normal and delta inference provide the exact same result at any time step. To increase the sparsity in $\Delta O(t)$, previous works introduced a threshold on the minimum amount of change to be propagated \citep{cbinfer} \citep{o2017sigma} \citep{neil2017delta}, which can result in a discrepancy between normal and delta inference (the latter being a rectified form of the former). The effect of this rectification may be seen as introducing noise, which a DNN should in principle be robust against. However, if this noise does not have a zero mean and since it is cumulative over time, it results in bias, which can lead to considerable drift; and  necessitates a periodical network reset or re-calibration to remediate. 

\subsection{Activation sparsification through quantization}

\begin{figure}[h!]
\begin{center}
\includegraphics[width=16cm]{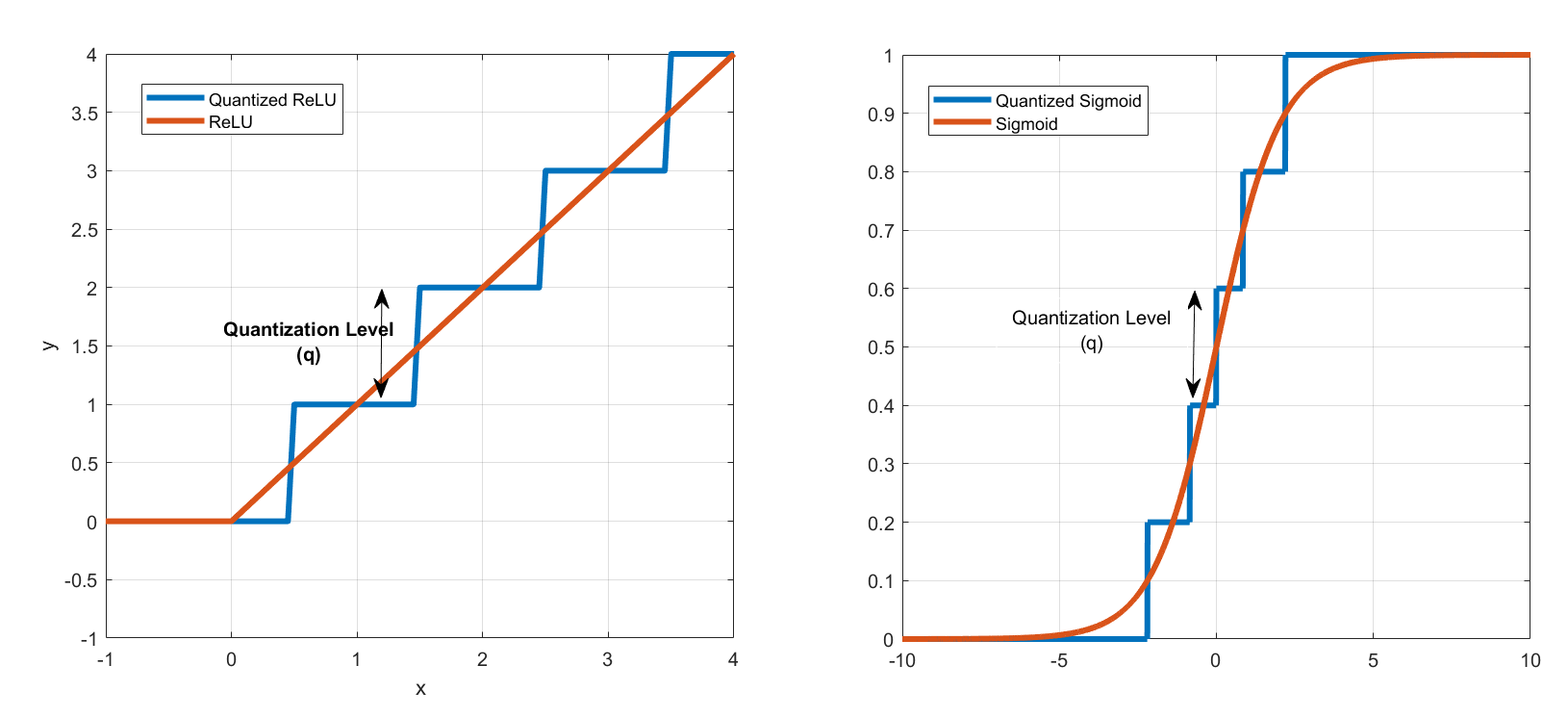} 
\end{center}
\caption{ReLU ($y=max(x,0)$), Quantized ReLU(with $q=1$), Sigmoid ($y=1/(1+e^{-x})$) and Quantized Sigmoid(with $q=0.2$) activation functions. Higher level of $q$ results in higher amount of sparsity in $\Delta O(t)$}
\label{fig:quantized_act_fun}
\end{figure}

Instead of a level-crossing threshold, here we propose a quantization method for the activations of a DNN, by replacing $f(.)$ with a quantized $f_q(.,q)$ in equations \ref{eq:activation_fun} and \ref{eq:delta_nn}; where $q$ refers to the quantization step-size.A larger quantization step-size decreases the resolution of the signal reconstruction levels (coarser quantization), and therefore neurons of the next layer receiving this activation signal will process less changes (deltas) between subsequent inputs at $(t-1)$ and $(t)$. Fig. \ref{fig:quantized_act_fun} shows an illustration of two quantized popular activation functions.


Note that the quantization operation of the Delta Activation Layer resembles the operation of a Sigma-Delta network \citep{o2017sigma}, however the Delta Activation Layer propagates a non-binary signal, and thus the herding-type of operations for the conversion to a binary spike-trains are spared. 
When using quantization (with a step-size) instead of a level-crossing threshold, \textbf{quantized inference and quantized delta-inference provide exact same results} and therefore no error will be accumulated over time (since the threshold on propagation of $\Delta O$ is zero). 

In addition, as we do quantization during training (in contrast with post-training threshold adjustment in \citep{cbinfer} and \citep{yousefzadeh2019asynchronous}), most of the accuracy drop due to the quantization error can be corrected with longer training.

\subsubsection{Temporal sparsity loss function}

By introducing the Delta Activation Layer, we need to add a regularization term in the loss function for encouraging sparsity optimization. A partial sum is added for every layer to the overall loss function and the sparsity factor ($\lambda$), is a hyper-parameter to adjust the contribution of sparsity loss of every layer in the total loss\footnote{Essentially}.

\begin{equation}\label{eq:L1} 
Loss_{Sparsity,l} =  \sum |\Delta O(t)_l|
\end{equation}

\begin{equation}\label{eq:total_loss} 
Loss_{Total} =  Loss_{Accuracy} + \sum_{l} \lambda_l \times Loss_{Sparsity,l}
\end{equation}

\noindent where $l$ is the layer index. 

Note that $Loss_{Sparsity,l}$ is entering the cost function as an L1-penalty term for a set of constraints that relate parameters though the activation tensors (and in this sense the sparsity factors $\lambda_l$ are the respective Lagrange multipliers). This implies that the $Loss_{Sparsity,l}$ and $Loss_{Accuracy}$ are competing and the optimization strives to find a set of parameters that balance the two losses for error (accuracy) and sparsity.

\subsection{Surrogate gradient of the quantization function}

As described the Delta\_activation layer framework and library takes advantage of activation quantization in order to introduce temporal sparsity. To recap, the way this is effected is as follows. Quantization confines and bins the real numbers into a fixed set centroids represented in the levels of the quantization scheme. When two numbers (e.g. two subsequent in time activations) lie nearby in range, namely within one quantization step, they get binned in the same level by the quantizer and their delta produces a zero. The smaller the set of levels (or the larger the quantization step) the more likely it gets for two numbers to get quantized in the same level (i.e. have the same centroid) and their delta to be zero. The smaller the set of levels, on the other hand, the larger the quantization error/noise introduced, which, when it is in the same range as the optimization error, the training convergence can slow down or get stuck prematurely, and consequences for the final achievable model accuracy. There is therefore an objective tussle and an optimal trade-off between quantization induced sparsity and accuracy, which we would like optimization process to auto-fine-tune. And this motivates making the quantization step a ``learnable'' parameter in the cost function. This, also enables the additional flexibility of maintaining a different quantization step per-layer, or per-channel as opposed to a globally uniform one (with entails benefits in terms of resource utilisation and better sparsity control).

Now, in practise a quantization function is by virtue non-smooth and non-continuous, making its presence inside the cost function problematic in face of gradient-based optimisation (through back-propagation), simply because it has no closed-form gradient (it is non differentiable). For this reason one need to resort in an approximate or a surrogate gradient as a replacement.

The key insight for finding a good choice of a surrogate gradient to use with the Delta\_activation layer, lies in expressing mathematically the relationship between the quantization step (or number of levels) and the induced temporal sparsity by means of the delta operator (described above). And the goal is to formalize this relationship through a constraint, which can be made part of the cost function as a Lagrangian term; essentially the penalty term that we have shown in Eq. \ref{eq:L1} and \ref{eq:total_loss} above.

Here, we introduce and motivate the surrogate that we chose for our experiments, although for the herein described methodology this is neither the only plausible, nor likely the best possible one can use in the Delta\_activation layer.

In this work, we chose to perform quantization by using the straight-forward uniform (linear step) quantization function, shown in Eq.\ref{eq:quantized_fun}.

\begin{equation} \label{eq:quantized_fun}
f_q(Z,q) = round(\dfrac{f(Z)}{q}) \times q
\end{equation}

For gradient optimization with back-propagation, the gradient of $f_q(Z,q)$ with regard to $Z$ and $q$ needs to be computed for the minimization of the cost function of the accuracy loss and sparsity loss. As the quantization function is non-differentiable, we need to provide replacements for the actual gradients.

For the gradient of $f_q(Z,q)$ with regard to $Z$ we used the gradient of the non-quantized activation function, as it is typical with quantization training of neural networks \citep{jacob2018quantization}, and which dissolves to the straight-through estimator (STE) for $f(Z)$.

\begin{equation}\label{eq:accuracy_loss_grad} 
\dfrac{\partial f_q(Z,q)}{\partial Z} = \dfrac{\partial f(Z)}{\partial Z}
\end{equation}

Next, as we want the step-size $q$ in the Delta Activation Layer to also be a trainable parameter\footnote{The step-size ($q$) can be global for the entire network, or different per layer, or different per neuron. In addition, in the case of a CNN (as in our experiments), it can be shared across every channel (convolutional feature map), so as to ``interact'' with the bias term during the optimization process. In this case, the number of additional trainable parameters (quantization levels) equals the number of biases.}, the gradient of $f_q(Z,q)$ with regard to $q$ needs to be functionally connected to the credit assignment aspect of the error back propagation process, so as to influence with the right incentive the optimization process; namely towards increasing the quantization step $q$ and thereby the activation sparsity after the delta operation.

The quantization step $q$ gets updated with Eq.\ref{eq:q_update_grad}.

\begin{equation}\label{eq:q_update_grad} 
\Delta q = -\eta \dfrac{\partial Loss_{total}}{\partial q} \implies
\Delta q = -\eta ( \dfrac{\partial Loss_{accuracy}}{\partial q} + \sum_l \dfrac{\partial Loss_{sparsity,l}}{\partial q} )
\end{equation}

Because increase in $q$ generally results in decrease of the sparsity penalty $Loss_{sparsity}$, in its simplest probably form it suffices to take the surrogate of $\dfrac{\partial Loss_{sparsity,l}}{\partial q}$ to be equal $-|Loss_{sparsity,l}|$ or simply $-Loss_{sparsity,l}$ (since from Eq. \ref{eq:total_loss} $Loss_{sparsity,l}$ is an $L1$-norm and thus always positive).

\begin{equation}\label{eq:q_update_grad2} 
\dfrac{\partial Loss_{sparsity,l}}{\partial q} = -\eta Loss_{sparsity,l}
\end{equation}

Eq.\ref{eq:q_update_grad2} encourages higher increase in the quantization step $q$, for bigger $Loss_{sparsity,l}$. In turn the larger quantization step, leads to fewer quantization levels, and a higher probability for delta activations to result in zeros, thereby reduction in $Loss_{sparsity,l}$.

While this seemed to be consistent in our experimentation, we found that a further extension of it, where we use the current step size $q$ as a normalizing reciprocal (E.q. \ref{eq:sparsity_loss_grad}), gives an ``accelerated'' gradient and better results.

\begin{equation}\label{eq:sparsity_loss_grad}
\dfrac{\partial Loss_{sparsity,l}}{\partial q} = -\dfrac{Loss_{sparsity,l}}{q}
\end{equation}

\begin{equation}\label{eq:sparsity_loss_grad2} 
\Delta q = -\eta ( \dfrac{\partial Loss_{accuracy}}{\partial q} - \sum_l \dfrac{Loss_{sparsity,l}}{q} )
\end{equation}

The intuition now is that, when the penalty $Loss_{sparsity,l}$ is high, a small currently quantization step $q$ will be subjected to a larger increase than an already large quantization step. And when the penalty $Loss_{sparsity,l}$ is small (and the delta activation model is already sparse), a small currently quantization step will be subject to a rather small increase, while an already large quantization step will virtually remain the unchanged.

We can break-down the LHS and RHS of Eq. \ref{eq:sparsity_loss_grad} in components of the chain-rule as follows\footnote{Given $\dfrac{\partial|x|}{\partial x} = \dfrac{|x|}{x},  \quad x \neq 0$}:

\begin{equation} \label{eq:LHS}
LHS(Eq.\ref{eq:sparsity_loss_grad}):  \dfrac{\partial Loss_{sparsity,1}}{\partial q} = \sum{\dfrac{\partial|\Delta O_l|}{\partial \Delta O_l} \times \dfrac{\partial\Delta O_l}{\partial q}} = \sum{\dfrac{|\Delta O_l|}{\Delta O_l} \times \dfrac{\partial\Delta O_l}{\partial q}}
\end{equation}

\begin{equation} \label{eq:RHS}
RHS(Eq.\ref{eq:sparsity_loss_grad}):  \dfrac{Loss_{Sparsity,l}}{q} =  \sum \dfrac{|\Delta O_l|}{q} = \sum \dfrac{|\Delta O_l|}{\Delta O_l} \times \dfrac{\Delta O_l}{q}
\end{equation}

\noindent Comparing Eq.\ref{eq:LHS} and Eq.\ref{eq:RHS} we can conclude the final form of the surrogate of the gradient with regard to the quantization step (which we used in Keras):

\begin{equation}\label{eq:surrogate_gradq}
\dfrac{\partial \Delta O_l}{\partial q} = - \dfrac{\Delta O_l}{q} \implies
\dfrac{\partial O_l}{\partial q} = - \dfrac{O_l}{q} \implies \dfrac{\partial f_q(Z,q)}{\partial q} = -\dfrac{f_q(Z,q)}{q}
\end{equation}

So finally, substituting Eq. \ref{eq:surrogate_gradq} back in Eq. \ref{eq:q_update_grad} we get 

\begin{equation} 
\begin{aligned}
\Delta q = -\eta \, \dfrac{\partial Loss_{total}}{\partial q}
\implies
\Delta q = -\eta ( \dfrac{\partial Loss_{accuracy}}{\partial q} + \sum_l \dfrac{\partial Loss_{sparsity,l}}{\partial q} ) 
\\
\implies
\Delta q = -\eta \, (
\dfrac{\partial Loss_{accuracy}}{\partial f_q(Z,q)} \times
\dfrac{\partial f_q(Z,q)}{\partial q} + 
\sum_l \dfrac{\partial Loss_{sparsity,l}}{\partial f_q(Z,q)} \times
\dfrac{\partial f_q(Z,q)}{\partial q}
)
\\
\implies
\Delta q = -\eta \, (
\dfrac{\partial Loss_{accuracy}}{\partial f_q(Z,q)} \times
\dfrac{ -f_q(Z,q)}{ q} + 
\sum_l \dfrac{\partial Loss_{sparsity,l}}{\partial f_q(Z,q)} \times
\dfrac{ -f_q(Z,q)}{ q}
)
\\
\implies
\Delta q = \eta \, \dfrac{ f_q(Z,q)}{ q} \, (
\dfrac{\partial Loss_{accuracy}}{\partial f_q(Z,q)}  + 
\sum_l \dfrac{\partial Loss_{sparsity,l}}{\partial f_q(Z,q)} 
)
\end{aligned}
\end{equation} 

To summarize our answer, the surrogate gradients are as follows:

\begin{equation}
\dfrac{\partial f_q(Z,q)}{\partial Z} = \dfrac{\partial f(Z)}{\partial Z}
\end{equation}

\begin{equation} 
\dfrac{\partial f_q(Z,q)}{\partial q} = -\dfrac{f_q(Z,q)}{q}
\end{equation}

It is also worth pointing out that there are several options \citep{sparseDNN2019} that can be used to encourage sparsity of the $\Delta O$ activations tensor, most of which are compatible with the Delta Activation Layer. The more promising options in recent literature \citep{kurtz2020inducing} are the various forms of the Hoyer penalty, and the difference of $L1$ and $L2$ norms. Nevertheless, as our focus is in show-casing the temporal sparsification pipeline with the Delta Activation Layer, we confined our experimentation with the more popular $L1$ penalty, for comparison purposes against a larger corpus of DNN literature, e.g. \citep{georgiadis2019accelerating}.

\subsection{Selective spatio-temporal sparsification}

\begin{figure}[h!]
\begin{center}
\includegraphics[width=15cm]{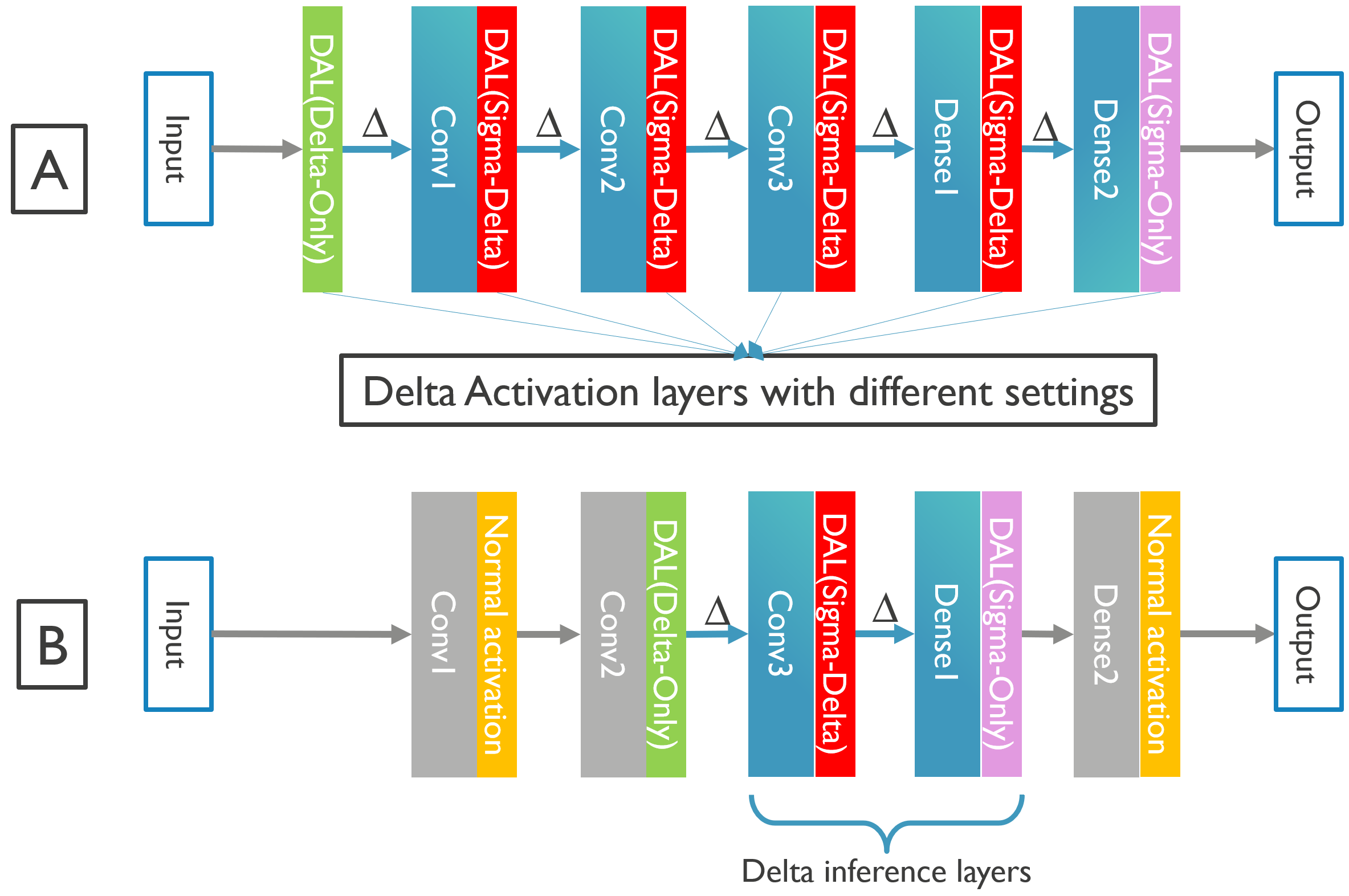} 
\end{center}
\caption{A) An example of a delta inference when the input and output of the network are not delta type. DAL stands for Delta Activation Layer. DAL(Delta-Only) layer is Delta Activation Layer where the integration phase is skipped. Similarly, DAL(Sigma-Only) layer is a configuration of the Delta Activation Layer where the differentiation is skipped. And DAL(Sigma-Delta) layer is a complete Delta Activation Layer. B) When only 2 layers of the 5-layer DNN is configured to process in the delta inference method. Normal activation layers are activation layers without state and therefore are cheaper to be implemented. Normal activation may also use L1 regularization on the activations to improve spatial sparsity as described in \citep{kurtz2020inducing}}
\label{fig:partial_delta}
\end{figure}

Since delta type of inference incurs additional operations and memory overheads per each non-zero delta value, one may prefer to only use it for specific layers where the saving from sparsity can be considerable. Typically, besides the linear operations in Eq. \ref{eq:layer}, two extra operations (integration in Eq. \ref{eq:sigma_nn} and differentiation in Eq. \ref{eq:delta_nn}) are required. However, if one wants to feed-in a Delta Activation Layer from a ``normal'' activation layer, as the input will be $X(t)$ instead of $\Delta X(t)$, the integration step in Eq. \ref{eq:sigma_nn} should be skipped. Similarly, if the output of a Delta Activation Layer is connected to a ``normal'' inference layer, the differentiation operation in Eq. \ref{eq:delta_nn} should be skipped. Two figurative examples of such configurations are laid out in Fig. \ref{fig:partial_delta}.

\subsection{Delta Activation Layer configurations}

Our Delta Activation Layer comes with many several configuration options to make the use customizable. Here is a list of most important parameters:

\begin{enumerate}
    \item Quantization mode: the granularity of applying quantization, which can be neuron-wise, channel-wise or layer-wise

    \item Activation function: the type of activation function for DNN layer (e.g ReLU, Softmax, etc.)
        
    \item Sparsity factor ($\lambda$ in E.q. \ref{eq:total_loss})
    
    
    \item Max Pooling: Poolings are one of the most important operations in the DNN. Even-tough some types of poolings are linear (for example Average Pooling or Stride) and can be integrated normally as part of Eq. \ref{eq:layer}, Max Pooling is an exception. It is not possible to apply Max Pooling directly on $\Delta O(t)$, since it is a non-linear operation. When Max Pooling is used, it should be applied on $O(t)$. Therefore it is implemented as an option inside the Delta Activation Layer. To use Max\_Pooling, this option should be activated and the size of the pooling should be defined. 
    
    \item Integration skip: if the integration operation (Eq.\ref{eq:sigma_nn}) should be skipped or not.

    \item Differentiation skip: if the differentiation operation (Eq.\ref{eq:delta_nn}) should be skipped or not. 
\end{enumerate}

We have used channel-wise quantization mode in all experiments in this paper, since it is a good balance between number of parameters and performance, and suits the CNN structure. 


The optimal selection of a sparsity factor may be automated through grid-searched as a hyper-parameter. However, in practise it is subject to user/system requirements since it is the knob that controls the trade-off between activation sparsity rate\footnote{Activation \textit{sparsity rate} which measures the rate of zero activations in a feature map, is not to be confused with \textit{sparsity factor} $\lambda_l$ in E.q. \ref{eq:total_loss}; albeit the two are correlated by virtue of optimizing the cost function.} (compute and power efficiency) and accuracy. For the experiments in this work we have used a heuristic way of setting the sparsity factor relating to each Delta Activation Layer such that it is proportional to the the fan-out in terms of MAC operations downstream. For example when a layer is connected to 128 convolution filters with $3 \times 3$ kernels, every non-zero neuron activation feeds in $128\times 3\times 3$ MAC operations. Since our final goal is to reduce the number of operations, we increased the sparsity factor more for layers whose neurons can triggers more MAC operations.



\section{Results} \label{sec:results}



Our experiments with temporally sparse video inference are mostly based on the ResNet-50 \citep{resnet} and MobileNet \citep{howard2017mobilenets} DNN architectures, and the challenging UCF-101 dataset. UCF-101 contains 13320 videos from 101 action categories for human action recognition taken with both fixed and moving cameras. Even though a fixed camera seems more appealing for exploiting temporal sparsity, our results show that temporal sparsity in deeper layers of a DNN is also very high in recordings with a moving camera.

ResNet-50 and MobileNet were originally used for image processing, however we used them here for video stream processing by employing it in the \textit{single-frame} arrangement described in \citep{google2014large}, and leaving the temporal dimension as the task for the delta inference processing. Having said that the best accuracies for the UCF101 dataset to our knowledge have been achieved using 3D convnets with two-stream input (frames and optical flows)\citep{carreira2017quo}. The simpler model we consider however, permits an easier understanding of the sparsification effects of the Delta Activation Layer.

To benchmark the sparsity/accuracy trade-off when applying the Delta Activation Layer we have used 4 different setups\footnote{training/fine-tuning were performed on an NVIDIA Geforce RTX 2080, but experiments are expected to be reproducible in any recent CUDA-enabled GPU.}. In the \textbf{baseline} setup we have fine-tuned Resnet-50 for UCF101 dataset without using any sparsity penalty. In this case, we used a pre-trained ResNet-50 for `image-net' dataset \citep{resnet} and fine-tune only its last Dense (fully connected) layer and the output layer for UCF101 dataset. We used a standard input image size of $(224,224)$, batch size of $32$, and the SGD optimizer for $8$ epochs. We have tried to extend fine-tuning to more layers but the best results were achieved by fine-tuning the last two layers. For MobileNet, we fine-tuned the whole network for $3$ epochs.

For our second setup, we applied ``L1 regularization loss" on activations of the baseline network to have a conventional spatially sparse DNN. For this setup (and the remaining two setups), we had to fine-tune (re-train) the whole ResNet-50 network. For \textbf{Spatial Sparsification} setup, we started from the pre-trained weights of our baseline network and re-train all the layers until it is relatively converged.

\begin{figure}[h!]
\begin{center}
\includegraphics[width=15cm]{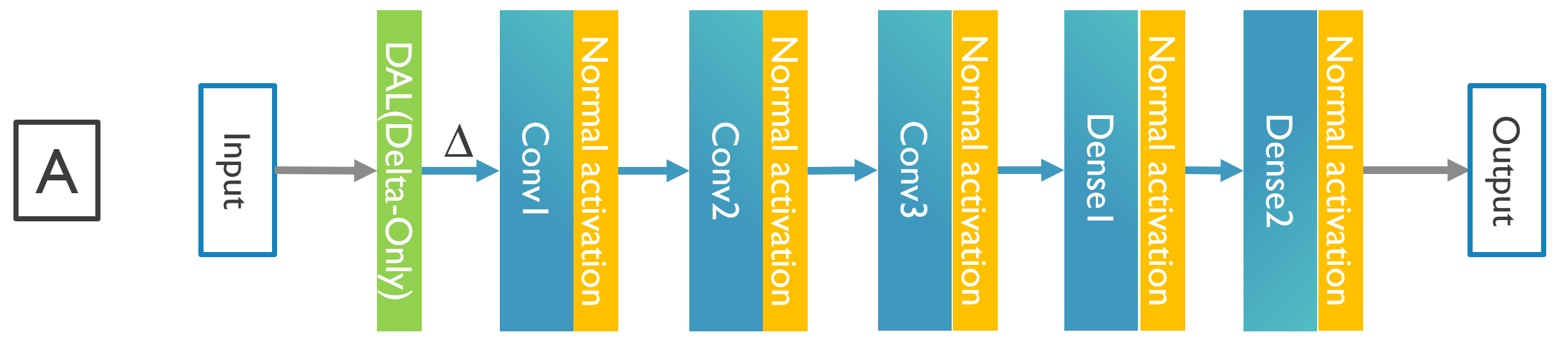} 
\end{center}
\caption{A) An example of a delta inference when only input is in the form of delta \citep{PIX2NVS}. This network exploits temporal sparsity only in its input (DAL stands for Delta Activation Layer). Here we skipped the `DAL(Sigma-Only)' layer and process the input deltas directly in the following layers. In this case, the following layers process the deltas without remembering the past. This network should learn to recognize actions only based on the delta frames and won't be equivalent to the original DNN. please note that the number of layers and network configuration in this figure is only an example and does not represent ResNet-50.}
\label{fig:input_delta}
\end{figure}

As the third setup for our experiments, we have used a network where a Delta Activation Layer is only used as the input layer as illustrated in Fig. \ref{fig:input_delta}. In this case, the network only processes changes of the scene (deltas only) without having the reference background in its state, which is sparse, to begin with. This is equivalent to providing inputs from a DVS sensor \citep{DVS_tobi}, or similar to the method presented in \citep{PIX2NVS} where the input deltas are considered as a simpler version of optical flows. The ``normal activation" layers in this experiment are also equipped with ``L1 regularization" of activations during training for higher spatial sparsity. Therefore this setup is called \textbf{Spatial sparsification + Input delta}. Same as the second setup, here we started from a baseline network and fine-tune all the layers until it relatively converges.

The last setup is using Delta Activation Layer in place of all the activation layers in ResNet-50 network (same as Fig. \ref{fig:partial_delta}(A)) and therefore results in exploitation of temporal sparsity in all the layers of ResNet-50. We called this setup \textbf{Temporal sparsification}. Again, here we started from the pre-trained weights of the baseline model and fine-tune all the layers until a relatively good convergence. In this setup and during training, Delta Activation Layer calculates temporal sparsity loss by using Eq.\ref{eq:L1} and perform activation quantization during training to minimize the temporal sparsity loss. We repeated the same process for MobileNet.

Table \ref{tb:1_results} reports results on sparsity gains during inference versus accuracy, from experiments with different training setups of the ResNet-50/MobileNet networks (alongside some reference accuracy scores achieved in the literature for the same dataset, albeit by using different DNN architectures).

\begin{table}
\centering
\begin{tabular}{|c|c|c|}
 \hline
 \textbf{Training setup}                 & \textbf{Classification accuracy }  & \textbf{Operation Sparsity}    \\ 
 \hline
 3D two-stream CNN \citep{kalfaoglu2020late}    & 98.69\%    & --            \\
 \hline
 Slow fusion in \citep{google2014large}         & 65.4\%                & --            \\
 \hline
 Spatial convent in \citep{two_stream_2014}     & 72.8\%                & --            \\
 \hline
 \hline
 ResNet-50 (Baseline)                           & 73\%                  & 49.1\%        \\
 \hline
 ResNet-50 (Spatial sparsification)             & 65.4\%                & 65.3\%         \\
 \hline
 ResNet-50 (Spatial sparsification + Input delta)   & 70.4\%            & 68.8\%        \\
 \hline
 \textbf{ResNet-50 (Temporal sparsification)}   & \textbf{67.6\%}               & \textbf{93.1\%} \\
 \hline
 \hline
 MobileNet (Baseline)       & 77.1\%                & 38.2\%        \\
 \hline
 \textbf{MobileNet (Temporal sparsification)}   & \textbf{73\%}               & \textbf{77.7\%} \\
 \hline
\end{tabular}
 \caption{Results for UCF101-based human action recognition, from our experiments and some those reported in previous works. The reported results from \citep{google2014large} and \citep{two_stream_2014} have used 2D convolutions over individual frames (similar to our approach), while \citep{kalfaoglu2020late} used a more complex approach and we include their results here only to reference the best achieved accuracy on this dataset. In our experiments, the classification accuracy is the average of the classification results across the 32 consecutive frames (for every each video in the dataset). Operation sparsity is the ratio of MAC operations with zero activations over the total number of MAC operations (weight sparsity is not accounted here).}
 \label{tb:1_results}
\end{table}

The results reported in Table \ref{tb:1_results} aim to highlight a couple of important aspects. First of all similarly to spatial sparsification (e.g. by means of L1 Regularization), temporal sparsification my means of the Delta Activation Layer may affect adversely the accuracy. Since adding the Delta Activation Layer requires quantization and fine-tuning of the entire network (in comparison with the baseline network where only the last layers are fine-tuned), there is a tussle between training error and quantization error that can either lower accuracy or cost the training process more iterations to recover it\footnote{we noticed that different quantization schemes seem to affect variably the percent of accuracy loss in conjunction with the time the model is allowed to train. The longer the training time, the closer it comes to recovering the accuracy loss, but at a slowing-down convergence rate (which also gets even slower with coarser quantization levels). Due to the time constraints the stopping criterion for which we reported the results was based on number of epochs and not the convergence error.}. Second, at the same time, the gain in sparsity increase is also considerable when accounting for temporal sparsification. Table \ref{tb:1_results} shows that spatio-temporal specification results in better sparsity in the operations while the accuracy drop is in the same range of ``L1 Regularization" for spatial sparsity.

\begin{figure}[h!]
\includegraphics[width=18cm]{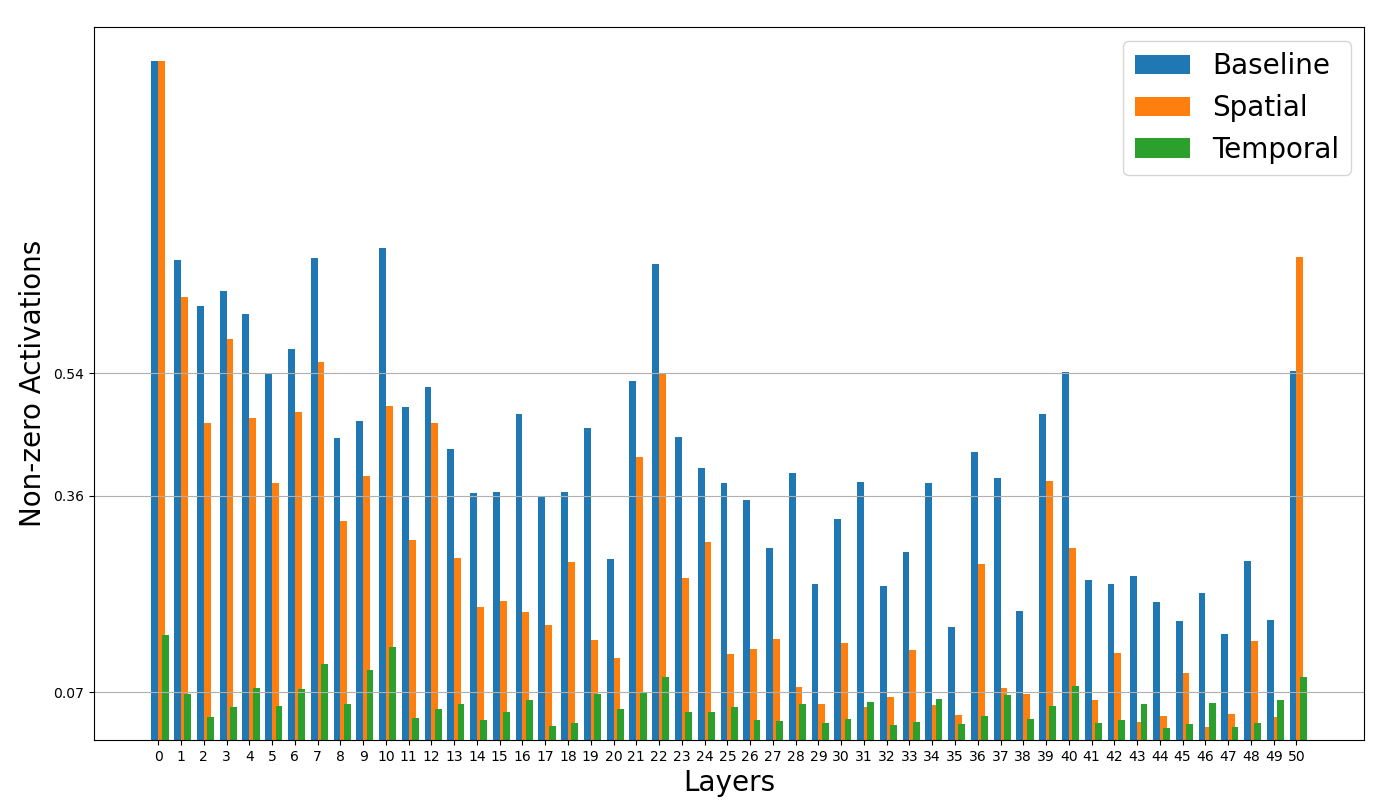} 
\caption{The number of non-zero activations for the baseline network, spatial sparsification with ``L1 regularization", and temporal sparsification with Delta\_Action layer in Resnet-50. Layer `0' is the input layer and the Layer `50' is the dense layer. The output layer is not shown since activations in the output layer does not trigger any internal operation. Layer indexing here is always the same as original ResNet-50. Delta Activation Layers are not considered as extra layers but as a replacement for activations in the original network.}
\label{fig:sp-layer-wise}
\end{figure}

\begin{figure}[h!]
\begin{center}
\includegraphics[width=10cm]{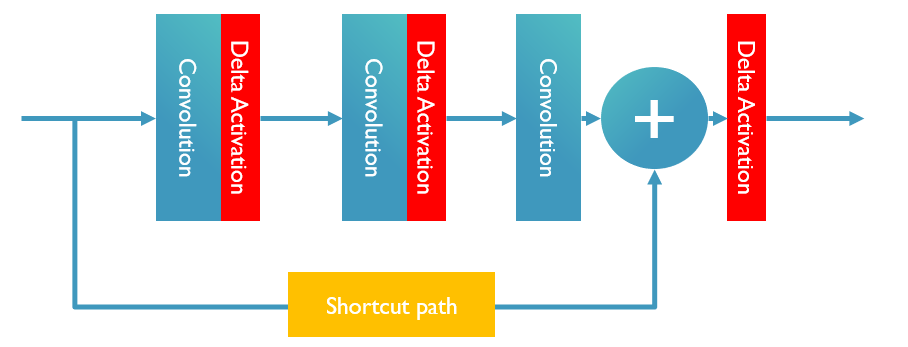} 
\end{center}
\caption{Residual block in ResNet-50, the shortcut path is either just a bypass connection or it contains a convolution operation. Temporal sparsity in the Delta Activation Layer just after the ADD operation is minimum.}
\label{fig:Resnet_block}
\end{figure}

Fig. \ref{fig:sp-layer-wise} shows details ratio of activation sparsity in ResNet-50 per layer. It is evident that both temporal and spatial sparsity in general increase when going deeper inside the ResNet-50 network. A fully connected layer (layer 50) has less spatio-temporal sparsity than previous layers. It is interesting to note that the layers where skip connections are merging with the main branch (layers 4,7,10,13,16,..,46), illustrated in Fig.\ref{fig:Resnet_block}, exhibit the lowest temporal sparsity in their neighbourhood (non-zero deltas are maximum compares to the neighboring layers), while this is not the case for spatial sparsity.

\begin{figure}[h!]
\begin{center}
\includegraphics[width=10cm]{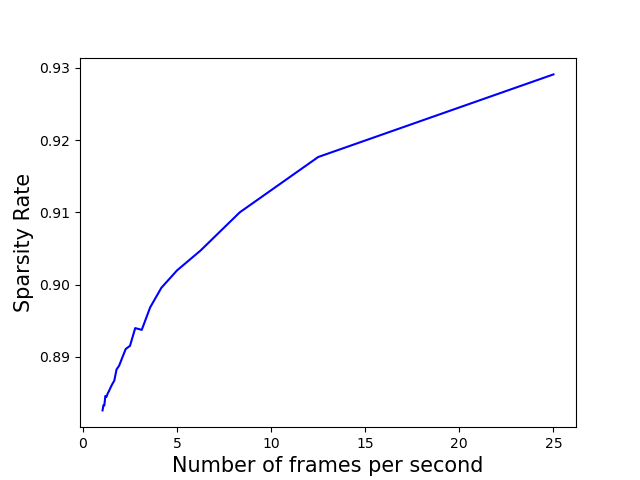} 
\end{center}
\caption{Activation sparsity rate (percent of zero activations in a feature map) versus number of frames per second}
\label{fig:sparsity_fps}
\end{figure}

One observed advantage of temporal sparsity in these experiments, is that number of inference operations does not scale up linearly with the increase in frame rate. Increasing the frame-rate increases the temporal precision of the network and is important when a low-latency inference is required (for example in self-driving cars). However, increasing the frame-rate does not increase the amount of computations linearly. In general, higher frame-rates results in higher temporal sparsity as our results in Fig. \ref{fig:sparsity_fps} show; where we run the inference on UCF101 dataset with different frame-rates (by sub-sampling in time). This observation is more prominent with higher frame rates (like 120fps).

\begin{figure}[h!]
\includegraphics[width=18cm]{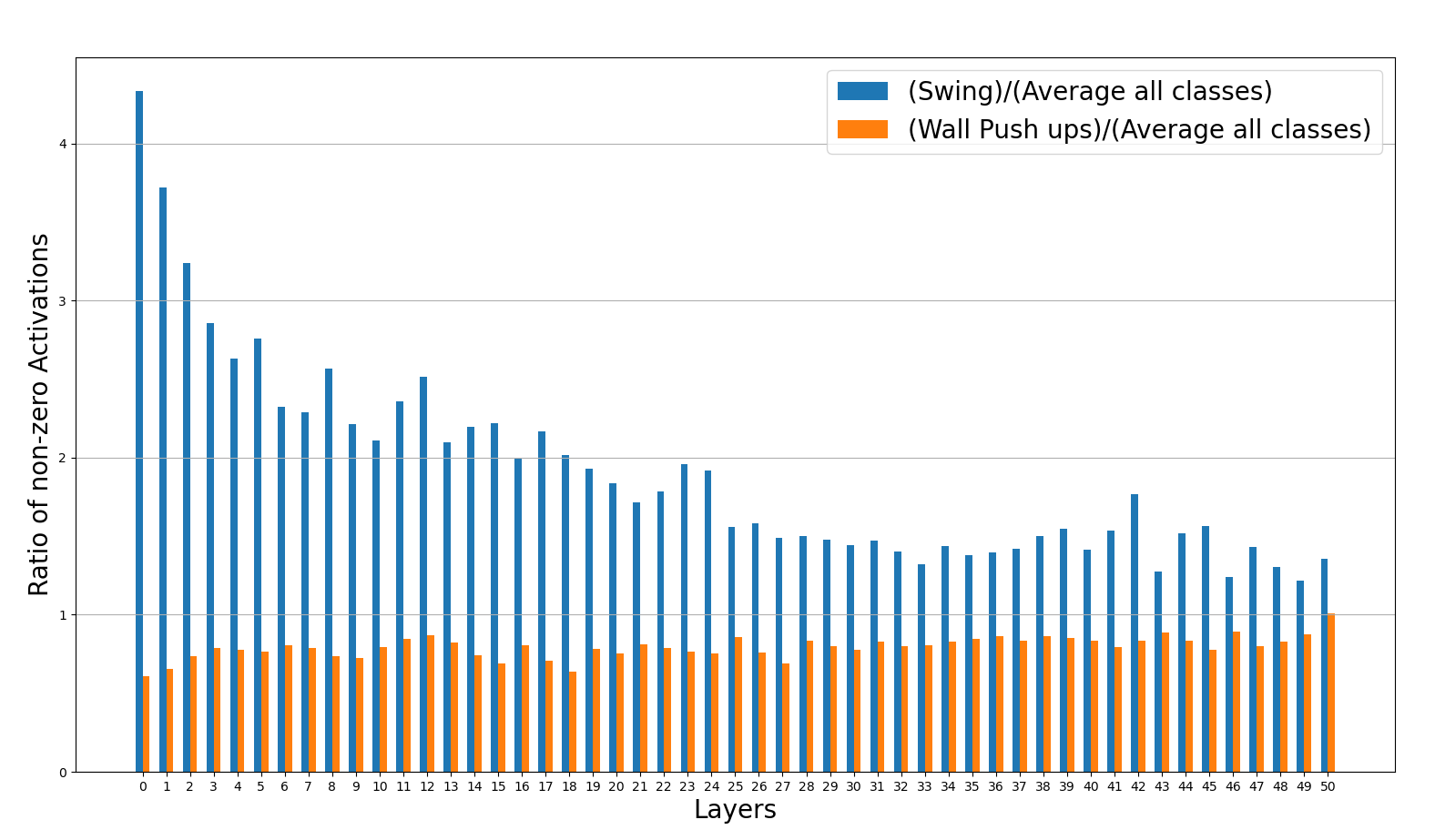} 
\caption{Average ratio of non-zero activations (deltas) for videos of `swing' action (lowest temporal sparsity) and videos of `wall push ups' (highest temporal sparsity) versus the average sparsities of all classes. It is interesting to see that the low/high amount of changes in the input frame mostly affect the low level features in the beginning layers.}
\label{fig:swing_wallpushup}
\end{figure}

The degree of temporal sparsity is also dependent on the type of motion actions in the scene. In Table \ref{tb:sparsity_classes} we list the `3' classes with highest average temporal sparsity across layers and the `3' classes with lowest. For the low sparsity classes, we noticed that most of the videos are recorded with moving cameras while the highly sparsity classes correspond to videos recorded with fixed cameras. Also, the actions in videos of high sparsity classes are slow and limited. We therefore were curious to find out how the amount of movement in the input affects the temporal sparsity in the features generated deeper inside the neural network. Fig. \ref{fig:swing_wallpushup} shows the layer-wise non-zero deltas for two classes of the test set, the one with lowest temporal sparsity (swing) and the one with highest temporal sparsity (wall push ups), normalized by the average sparsity degree of all classes. In general, the effect of movement in the input frames has less effect on the last layers of neural network where high level features (concepts) are extracted.

\begin{table}
\centering
\begin{tabular}{ |c|c|}
 \hline
 Class name   &    Activation Sparsity \\
 \hline
 Swing        &    83.5\%               \\
 \hline
 Punch        &    84.2\%               \\
 \hline
 Lunges       &    84.5\%               \\
 \hline
 Cricket Bowling &    93.2\%            \\
 \hline
 Writing On Board        &    93.9\%     \\
 \hline
 Wall Push ups        &    94.4\%         \\
 \hline
\end{tabular}
 \caption{Selected classes of UCF101 dataset with highest and lowest activation sparsities in temporally sparse ResNet-50.} 
 \label{tb:sparsity_classes}
\end{table}

\begin{figure}[h!]
\begin{center}
\includegraphics[width=10cm]{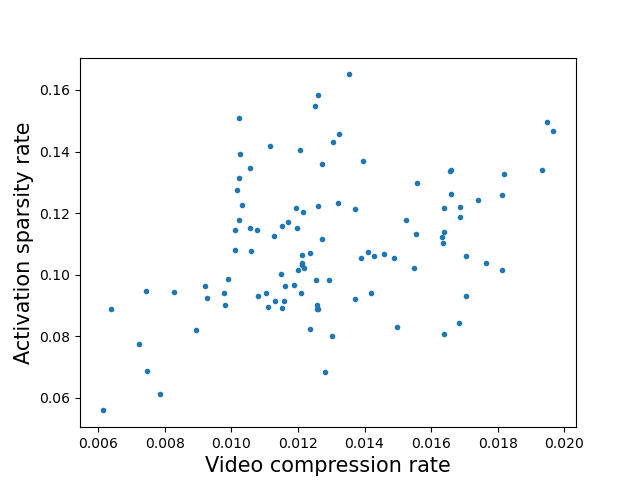} 
\end{center}
\caption{Averaged activation sparsity rate versus the averaged video compression rate for each UCF101 class (i.e. data points are centroids}
\label{fig:sparsity_compression}
\end{figure}

We also tried to find a correlation  between the degree of temporal sparsity and the rate of compression by the video codec used. Fig. \ref{fig:sparsity_compression} suggests that there maybe some weak correlation but this relationship is not straight-forward. Our calculated correlation coefficients using Pearson (assuming Gaussianity) and Spearman (non-Gaussianity assumptions) methods \citep{artusi2002bravais} give a coefficient of about $0.36\%$. PCA and ICA analyses point to another strong component in video compression rate beside temporal activation sparsity rate. Video compression techniques are very advanced compared to our simple temporal delta processing and while both temporal and spatial redundancies are properly exploited in the compression algorithms there are other also other aspects that such clever algorithms capture (e.g. non-uniform redundancies by learning dictionaries of sequences or filters). It can be seen that the compression rates in videos are an order of magnitude higher than the activation sparsity rates. This means there is still a large space for improvement to ``properly'' sparsify DNN computations.

\section{Discussion} \label{sec:discussion}


In this work, we introduced a method of adding temporal sparsity in general DNN inference on video data. There are two key aspects that make this work distinctive. One is the fact that given a trained model we look into sparsifying activations arbitrarily deep in the network structure, i.e. we do not only focus on preprocessing the inputs or the first layers of the network. Rather, the proposed mechanism is deployable after any layer, and can be flexibly employed with any subset of the network's layers. The other aspect is that our approach can work synergetically with training. This means that where refinement phase (e.g. transfer-learning) or vanilla training is feasible, an optimization constraint is introduced to leverage the sparsification effectiveness of our mechanism during inference.

\begin{figure}[h!]
\includegraphics[width=18cm]{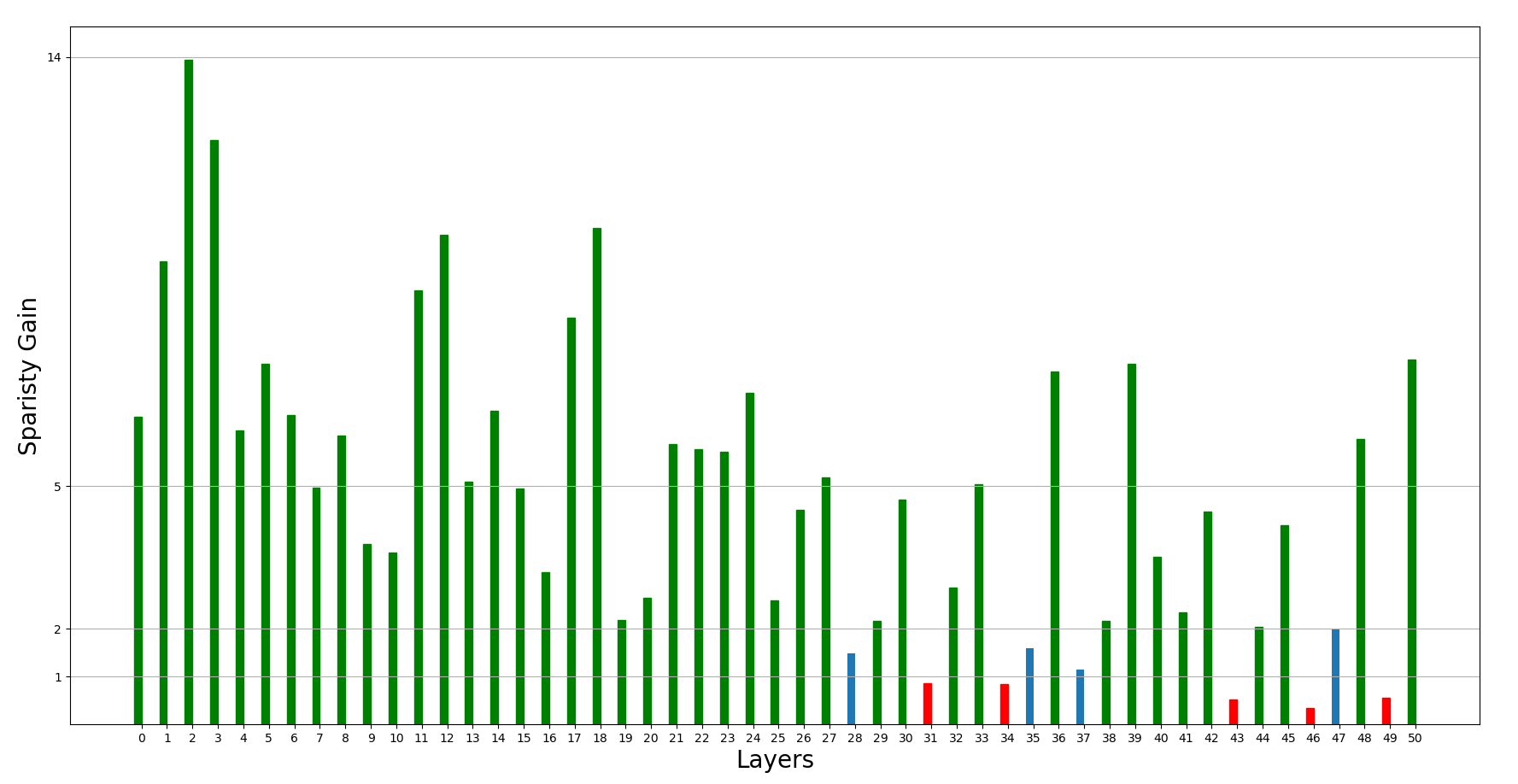} 
\caption{Layer-wise sparsity gain for temporal sparsity setup over spatial sparsity setup (ratio of non-zero activations) in our experiments with ResNet-50 and UCF101. The average gain is 5. The blue/red bars are the low gain layers with the gains of less than two/one.}
\label{fig:sp-layer-wise-gain}
\end{figure}

Integration of the Delta Activation Layer in a DNN is not cost-free. There are two main overheads introduced which need to be weighed against the benefits of introducing temporal sparsity (discussed below). For this reason we promote the selective use of the Delta Activation Layer only between layers where there is a clear gain to be made. More specifically,  in Fig.\ref{fig:sp-layer-wise-gain} we observe that the temporal sparsity gain (over spatial sparsity) is very different in different layers.

The first overhead introduced by a Delta Activation Layer stems from the requirement to keep track of the associated neuron states, which increases the overall memory footprint.
For example in ResNet-50 where the number of parameters is almost 23M, there are 9.2M neurons. When using normal inference (not using Delta Activation Layer), it is required to store only activations for two consecutive layers at any moment during inference (the input to a layer and its output). To account for the widest layers this is only about 1M neurons which leads to total 25MB of memory (when states store in 16b floating-point format and weights are 8b). Inference with full Delta Activation Layer in all layers requires about 41MB compared to 25MB for normal inference of a single frame. In general, memory footprint is a serious constraint in all DNN accelerators and for this reason newer DNN accelerator architectures are catering for it (e.g. use of flash-memory is becoming more mainstream \citep{mythic}). The new NVIDIA Ampere architecture contains 40GB of DRAM memory which is 70\% more than its predecessor).

The second overhead, and since more memory is involved, is memory access time and energy. Increasing memory footprint may force the DNN accelerator architecture to use a hierarchical memory structure to balance the memory footprint and the platform cost. For example, external DRAM memory is two orders of magnitude cheaper than a local SRAM memory but at the same time it consumes two orders of magnitude more energy \citep{efficient_DNN_survey}. To have a clear picture, imagine a DNN accelerator with only 1MB of on-chip memory. For the normal inference with ResNet-50, we could use the local memory for  activations state (read and write operations) and external memory for the weights (read operations). However, for delta-based inference, we are forced to use external memory for both weights and neuron activations. This will results in three times more external memory accesses (reading weights and states, writing states back) which is almost equal to three to five times more energy consumption (since external memory access consumes much more than any internal operation, and since neuron states normally have higher bit-width than weights). As a result, although we have observed that temporal sparsity typically reduces the number of operations, on average, by a factor of five or more when compared to spatial sparsification, in practice the amount of energy savings is less than 5x and very dependent on the memory hierarchy in the hardware. On the more optimistic side, however, our comparison against video compression rates in Fig. \ref{fig:sparsity_compression} suggests that there is plenty of room for achieving even higher sparsity, perhaps exploitable with more advanced algorithms, or future ramifications of the Delta Activation Layer. In addition, new memory technologies (like resistive RAM, embedded DRAM, and embedded Flash memory) may change the aforementioned cost calculations for on-chip memory in the near future.

Each Delta Activation Layer instantiation also introduces additional trainable parameters to the network. For our experiments with ResNet-50, a total of around 23K new parameters are added to the previous 23M weights and biases ($+0.1\%$). As the Delta Activation Layer introduces temporal sparsity constraint terms in the optimisation objective during training, these are in effect also regularizes against over-fitting. Besides that, it would also be possible to consider additional regularization terms in the cost function (e.g. Lasso, group Lasso, etc) targeting explicitly the weights/parameters controlling the network structure, rather than implicitly through the activations. In our experiment here we did not see such a need, however, a more systematic exploration and comparison with different regularisation regimes are the next steps for our future work with the Delta Activation Layer.

Other interesting aspects which we have not experimented with at this point, but which are in our plans for follow-up work include (a) to confirm the current results or their variability on different network architectures, but more importantly in relation to the capacity of the network, e.g. with lottery ticket \citep{frankle2018lottery} and distilled \citep{hinton2015distilling} type of networks; (b) the potential effects of other activation functions than ReLU in the effectiveness of the Delta Activation Layer, (c) using an advance quantization scheme along with the sparsity loss to reduce the accuracy drop and (d) the effect of lateral inhibition and winner-take-all strategies in promoting temporal sparsity \citep{ahmad2019can}.



\section*{Conflict of Interest Statement}

The authors declare that the research was conducted in the absence of any commercial or financial relationships that could be construed as a potential conflict of interest.

\section*{Author Contributions}

A.Y and M.S. contributed equally in this work. 

\section*{Funding}
This work was supported in part by the EU H2020 grants “TEMPO” and “ANDANTE”. 

\section*{Acknowledgments}

This work is partially funded by EU H2020 grants 826655 ``TEMPO” and 876925 ``ANDANTE".


\bibliographystyle{unsrtnat}
\bibliography{biblography}

\end{document}